\documentclass{article}

\usepackage{times}
\usepackage{graphicx} 
\usepackage{subcaption}

\usepackage{natbib}

\usepackage{algorithm}
\usepackage{algorithmic}

\usepackage{mathtools}
\usepackage{amsmath}
\usepackage{amssymb}
\usepackage{multirow}
\usepackage{tikz}
\usepackage{pstricks}
\usepackage{pgfplots}
\usepackage{caption,subcaption}

\usepackage{hyperref}

\usepackage[accepted]{icml2017}

\usepackage{lipsum}

\newcommand\blfootnote[1]{%
  \begingroup
  \renewcommand\thefootnote{}\footnote{#1}%
  \addtocounter{footnote}{-1}%
  \endgroup
}

\icmltitlerunning{Detecting Adversarial Samples from Artifacts}

\begin{document}

\twocolumn[
\icmltitle{Detecting Adversarial Samples from Artifacts}




\begin{icmlauthorlist}
\icmlauthor{Reuben Feinman}{caml}
\icmlauthor{Ryan R. Curtin}{caml}
\icmlauthor{Saurabh Shintre}{srl}
\icmlauthor{Andrew B. Gardner}{caml}
\end{icmlauthorlist}

\icmlaffiliation{caml}{Center for Advanced Machine Learning at Symantec, Mountain View, CA, USA}
\icmlaffiliation{srl}{Symantec Research Labs, Mountain View, CA, USA}

\icmlcorrespondingauthor{Reuben Feinman}{\texttt{reuben.feinman@nyu.edu}}
\icmlcorrespondingauthor{Ryan R. Curtin}{\texttt{ryan@ratml.org}}
\icmlcorrespondingauthor{Saurabh Shintre}{\texttt{saurabh\_shintre@symantec.com}}

\icmlkeywords{machine learning, neural networks, deep learning, adversarial,
uncertainty, bayesian inference, classification, undecided}

\vskip 0.3in
]



\printAffiliationsAndNotice{}  

\begin{abstract}
	Deep neural networks (DNNs) are powerful nonlinear architectures that are known 
to be robust to random perturbations of the input. However, these models are 
vulnerable to adversarial perturbations---small input changes crafted explicitly 
to fool the model. In this paper, we ask whether a DNN can distinguish adversarial 
samples from their normal and noisy counterparts. We investigate model confidence 
on adversarial samples by looking at Bayesian uncertainty 
estimates, available in dropout neural networks, and by performing density 
estimation in the subspace of deep features learned by the model. The result is 
a method for implicit adversarial detection that is oblivious to the attack 
algorithm. We evaluate this method on a variety of standard datasets including 
MNIST and CIFAR-10 and show that it generalizes well across different architectures 
and attacks. Our findings report that 85-93\% ROC-AUC can be achieved on a number 
of standard classification tasks with a negative class that consists of both 
normal and noisy samples.
\end{abstract}

\section{Introduction}
Deep neural networks (DNNs) are machine learning techniques that impose a hierarchical architecture consisting
of multiple layers of nonlinear processing units. In practice, DNNs achieve state-of-the-art
performance for a variety of generative and discriminative learning tasks from
domains including image processing, speech recognition, drug discovery and
genomics \cite{Lecun2015}.

\begin{figure}[t!]
  \vspace*{-1.5em}
  \begin{center}
  \includegraphics[width=0.4\textwidth]{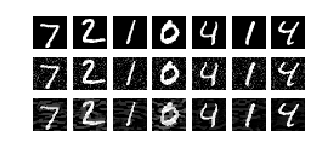}
  \end{center}
  \vspace*{-2.5em}
  \caption{Examples of normal (top), noisy (middle) and adversarial
(bottom) MNIST samples for a convnet. Adversarial samples were crafted via the
Basic Iterative Method \cite{Kurakin2017} and fool the model into misclassifying
100\% of the time.}
  \label{fig:image_examples}
  \vspace*{-0.7em}
\end{figure}

Although DNNs are known to be robust to noisy inputs \cite{Fawzi2016}, they have
been shown to be vulnerable to specially-crafted adversarial samples
\cite{Szegedy2014, Goodfellow2015}. These samples are constructed by taking a normal
sample and perturbing it, either at once or iteratively, in a direction that
maximizes the chance of misclassification. Figure \ref{fig:image_examples} shows
some examples of adversarial MNIST images alongside noisy images of equivalent
perturbation size. Adversarial attacks which require only small perturbations to
the original inputs can induce high-efficacy DNNs to misclassify at a high rate.
Some adversarial samples can also induce a DNN to output a specific target class
\cite{Papernot2016}. The vulnerability of DNNs to such adversarial attacks
highlights important security and performance implications for these models \cite{Papernot2016}.
Consequently, significant effort is ongoing to understand
and explain adversarial samples and to design defenses against them
\cite{Szegedy2014, Goodfellow2015, Papernot2016b, Tanay2016, Metzen2017}.
\blfootnote{The source code repository for this paper is located at \href{https://github.com/rfeinman/detecting-adversarial-samples}{http://github.com/rfeinman/detecting-adversarial-samples}}

Using the intuition that adversarial samples lie off the true data manifold,
we devise two novel features that can be used to detect adversarial samples:

\vspace*{-0.7em}
\begin{itemize} \itemsep -2pt
  \item \textbf{Density estimates}, calculated with the training set in the
feature space of the last hidden layer. These are meant to detect points that lie
far from the data manifold.

  \item \textbf{Bayesian uncertainty estimates}, available in dropout neural
networks. These are meant to detect when points lie in low-confidence regions of the input space,
and can detect adversarial samples in situations where density estimates cannot.
\end{itemize}
\vspace*{-0.7em}

When both of these features are used as inputs to a simple logistic regression
model, we observe effective detection of adversarial samples, achieving an ROC-AUC of
$92.6\%$ on the MNIST dataset with both noisy and normal samples as the negative
class. In Section \ref{sec:background} we provide the relevant background information for
our approach, and in Section \ref{sec:attacks} we briefly review a few state-of-the-art
adversarial attacks. Then, we introduce the intuition for our approach in Section
\ref{sec:manifold}, with a discussion of manifolds and Bayesian uncertainty.
This leads us to our results and conclusions in Sections
\ref{sec:experiments} and \ref{sec:conclusion}.
\section{Background}
\label{sec:background}

While neural networks are known to be robust to random noise \cite{Fawzi2016},
they have been shown to be vulnerable to adversarially-crafted perturbations
\cite{Szegedy2014, Goodfellow2015, Papernot2016}. Specifically, an
adversary can use information about the model to craft small perturbations that
fool the network into misclassifying their inputs. In the context of object
classification, these perturbations are often imperceptible to the human eye,
yet they can force the model to misclassify with high model confidence.

A number of works have attempted to explain the vulnerability of DNNs
to adversarial samples. \citet{Szegedy2014} offered a simple
preliminary explanation for the phenomenon, arguing that low-probability
adversarial ``pockets" are densely distributed in input space. As a result, they
argued, every point in image space is close to a vast number of adversarial
points and can be easily manipulated to achieve a desired model outcome.
\citet{Goodfellow2015} argued that it is a result of the linear nature of deep
classifiers. Although this explanation has been the most well-accepted in the
field, it was recently weakened by counterexamples \cite{Tanay2016}.
\citet{Tanay2016} introduced the `boundary tilting' perspective, suggesting
instead that adversarial samples lie in regions where the classification
boundary is close to the manifold of training data.

Research in adversarial attack defense generally falls within two categories:
first, methods for improving the robustness of classifiers to current attacks,
and second, methods for detecting adversarial samples in the wild.
\citet{Goodfellow2015} proposed augmenting the training loss function with an
additional adversarial term to improve the robustness of these models to a
specific adversarial attack. \textit{Defensive distillation} \cite{Papernot2016b} is another
recently-introduced technique which involves training a DNN with the softmax outputs of another neural
network that was trained on the training data, and can be seen as a way of
preventing the network from fitting too tightly to the data. Defensive
distillation is effective against the attack of \citet{Papernot2016}. However,
\citet{Carlini2016} showed that defensive distillation is easily broken with a
modified attack.

On the detection of adversarial samples, \citet{Metzen2017} proposed augmenting
a DNN with an additional ``detector'' subnetwork, trained
on normal and adversarial samples. Although the authors show compelling
performance results on a number of state-of-the-art adversarial attacks, one
major drawback is that the detector subnetwork must be trained on generated
adversarial samples. This implicitly trains the detector on a subset of all
possible adversarial attacks; we do not know how comprehensive this subset is,
and future attack modifications may be able to surmount the system.
The robustness of this technique to random noise is not currently known.
\section{Adversarial Attacks}
\label{sec:attacks}

The typical goal of an adversary is to craft a sample that looks similar to a
normal sample, and yet that gets misclassified by the target model. In the realm of
image classification, this amounts to finding a small perturbation that, when
added to a normal image, causes the target model to misclassify the sample, but
remains correctly classified by the human eye. For a given input image $x$, the
goal is to find a minimal perturbation $\eta$ such that the adversarial input
$\tilde{x} = x + \eta$ is misclassified. A significant number of adversarial attacks satisfying this goal have been
introduced in recent years.  This allows us a wide range of attacks to choose
from in our investigation.  Here, we introduce some of the most well-known and
most recent attacks.

{\bf Fast Gradient Sign Method (FGSM):} \citet{Goodfellow2015} introduced the
Fast Gradient Sign Method for crafting adversarial perturbations using the
derivative of the model's loss function with respect to the input feature
vector. Given a base input, the approach is to perturb each feature in the
direction of the gradient by magnitude $\epsilon$, where $\epsilon$ is a
parameter that determines perturbation size. For a model with loss $J(\Theta, x,
y)$, where $\Theta$ represents the model parameters, $x$ is the model input, and
$y$ is the label of $x$, the adversarial sample is generated as

\begin{equation*}
x^* = x + \epsilon\operatorname{sign}(\nabla_{x} J(\Theta, x, y)).
\end{equation*}

With small $\epsilon$, it is possible to fool DNNs trained for the MNIST and
CIFAR-10 classification tasks with high success rate \cite{Goodfellow2015}.

{\bf Basic Iterative Method (BIM):} \citet{Kurakin2017} proposed an iterative
version of FGSM called the Basic Iterative Method.  This is a
straightforward extension; instead of merely applying adversarial noise $\eta$
once with one parameter $\epsilon$, apply it many times iteratively with small
$\epsilon$.  This gives a recursive formula:

\vspace*{-1.5em}
\begin{eqnarray*}
x^*_0 &=& x, \\
x^*_i &=& \operatorname{clip}_{x, \epsilon}(x^*_{i - 1} + \epsilon
\operatorname{sign}(\nabla_{x^*_{i - 1}} J(\Theta, x^*_{i - 1}, y))).
\end{eqnarray*}
\vspace*{-1.5em}

Here, $\operatorname{clip}_{x, \epsilon}(\cdot)$ represents a clipping of the
values of the adversarial sample such that they are within an
$\epsilon$-neighborhood of the original sample $x$. This approach is convenient because it allows extra control over the attack.
For instance, one can control how far past the classification boundary a sample
is pushed: one can terminate the loop on the iteration when $x^*_i$ is first
misclassified, or add additional noise beyond that point.

The basic iterative method was shown to be typically more effective than the
FGSM attack on ImageNet images \cite{Kurakin2017}.

{\bf Jacobian-based Saliency Map Attack (JSMA):} \citet{Papernot2016} proposed 
a simple iterative method for \textit{targeted} misclassification. By 
exploiting the forward derivative of a DNN, one can find an
adversarial perturbation that will force the model to misclassify into a
specific \textit{target class}. For an input $x$ and a neural network $F$, the
output for class $j$ is denoted $F_j(x)$. To achieve a target class $t$,
$F_t(X)$ must be increased while the probabilities $F_j(X)$ of all other classes
$j \neq t$ decrease, until $t = \operatorname{arg\,max}_j F_j(X)$. This is
accomplished by exploiting the adversarial saliency map, which is defined as

\vspace*{-1.3em}
\begin{align*}
S(X,t)[i] = 
\begin{cases} 
0, \text{ if }\ \frac{\partial F_t(X)}{\partial X_i} < 0
\text{ or } \sum_{j \neq t} \frac{\partial F_j(X)}{\partial X_i} > 0 \\  
(\frac{\partial F_t(X)}{\partial X_i})|\sum_{j \neq t} \frac{\partial F_j(X)}{\partial X_i}|,
\text{ otherwise} 
\end{cases}
\end{align*}
\vspace*{-0.9em}

for an input feature $i$. Starting with a normal sample $x$, we
locate the pair of features $\{i, j\}$ that maximize $S(X,t)[i] + S(X,t)[j]$,
and perturb each feature by a constant offset $\epsilon$. This process is repeated
iteratively until the target misclassification is achieved. This method can effectively produce
MNIST samples that are correctly classified by human subjects but misclassified
into a specific target class by a DNN with high success rate.

{\bf Carlini \& Wagner (C\&W):} \citet{Carlini2016} recently introduced a
technique that is able to overcome defensive distillation. In fact, their
technique encompasses a range of attacks, all cast through the same optimization
framework. This results in three powerful attacks, each for a different
distance metric: an $L_2$ attack, an $L_0$ attack, and an $L_\infty$ attack.
For the $L_0$ attack, which we will consider in this paper, the perturbation
$\delta$ is defined in terms of an auxiliary variable $\omega$ as

\vspace*{-1em}
\begin{equation*}
\delta^*_i = \frac{1}{2} \left(\operatorname{tanh}(\omega_i + 1)\right) - x_i.
\end{equation*}
\vspace*{-1.3em}

Then, to find $\delta^*$ (an `unrestricted perturbation'), we optimize over
$\omega$:

\vspace*{-1.5em}
\begin{equation*}
\operatorname{min}_\omega \left\| \frac{1}{2} \left(\operatorname{tanh}(\omega) +
1\right) - x \right\|^2_2 + c f\left(\frac{1}{2} \operatorname{tanh}(\omega) +
1\right)
\end{equation*}
\vspace*{-1.2em}

\noindent where $f(\cdot)$ is an objective function based on the hinge loss:

\vspace*{-1.5em}
\begin{equation*}
f(x) = \operatorname{max}(\operatorname{max}\{Z(x)_i : i \ne t \} - Z(x)_t,
-\kappa).
\end{equation*}
\vspace*{-1.5em}

Here, $Z(x)_i$ is the pre-softmax output for class $i$, $t$ is the target
class, and $\kappa$ is a parameter that controls the confidence with which the
misclassification occurs.

Finally, to produce the adversarial sample $x^* = x + \delta$, we convert the
unrestricted perturbation $\delta^*$ to a restricted perturbation $\delta$, in
order to reduce the number of changed pixels.  By calculating the gradient
$\nabla f(x + \delta^*)$, we may identify those pixels $\delta^*_i$ with little
importance (small gradient values) and take $\delta_i = 0$; otherwise, for
larger gradient values we take $\delta_i = \delta^*_i$.  This allows an
effective attack with few modified pixels, thus helping keep the norm of
$\delta$ low.

These three attacks were shown to be particularly effective in comparison to
other attacks against networks trained with defensive distillation, achieving
adversarial sample generation success rates of 100\% where other techniques were
not able to top 1\%.

\section{Artifacts of Adversarial Samples} \label{sec:manifold}

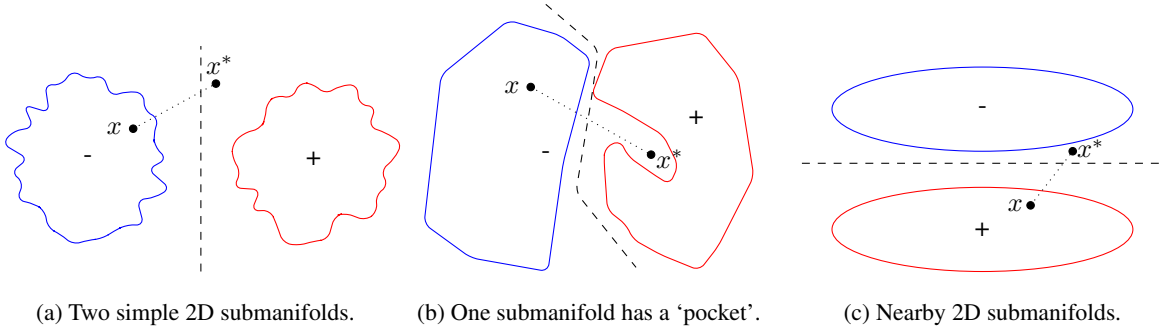
\begin{figure*}[htb]
\begin{center}
\begin{subfigure}[b]{0.3\textwidth}
\begin{center}
\newcommand\irregularcircle[2]{
  \pgfextra {\pgfmathsetmacro\len{(#1)+rand*(#2)}}+(0:\len pt)
  \foreach \a in {10,20,...,350}{
    \pgfextra {\pgfmathsetmacro\len{(#1)+rand*(#2)}} -- +(\a:\len pt)
  } -- cycle
}

\begin{tikzpicture}[scale=1.0]
  \draw [blue,rounded corners=1mm] (2,2) \irregularcircle{1cm}{2mm};
  \draw [red,rounded corners=1mm] (5,2) \irregularcircle{1cm}{2mm};
  \draw [style=dashed] (3.5, 0.5) -- (3.5, 3.5);
  \node [] at (2,2) { - };
  \node [] at (5,2) { + };

  \node [draw, circle, inner sep=1pt, fill] at (2.6, 2.4) { };
  \node [] at (2.35, 2.4) { $x$ };

  \node [draw, circle, inner sep=1pt, fill] at (3.7, 3.0) { };
  \node [] at (3.75, 3.25) { $x^*$ };

  \draw [style=dotted] (2.6, 2.4) -- (3.7, 3.0) { };
\end{tikzpicture}
\end{center}
\caption{Two simple 2D submanifolds.}
\label{fig:sit1}
\end{subfigure}
\begin{subfigure}[b]{0.3\textwidth}
\begin{center}
\newcommand\irregularthing[2]{
  \pgfextra {\pgfmathsetmacro\len{(#1)+rand*(#2)}}+(0:\len pt)
  \foreach \a in {10,20,...,350}{
    \pgfextra {\pgfmathsetmacro\len{abs(\a-180)/180*((#1)+rand*(#2))}} -- +(\a:\len pt)
  } -- cycle
}

\begin{tikzpicture}[scale=1.0]
  \draw [red,rounded corners=1mm]
      (5 - 0 + 0.1*rand, 0 + 0.1*rand) --
      (5 - 0.4 + 0.1*rand, 0.1 + 0.1*rand) --
      (5 - 0.8 + 0.1*rand, 0.3 + 0.1*rand) --
      (5 - 1.0 + 0.1*rand, 0.6 + 0.1*rand) --
      (5 - 1.1 + 0.1*rand, 0.8 + 0.1*rand) --
      (5 - 1.2 + 0.1*rand, 1.3 + 0.1*rand) --
      (5 - 1.3 + 0.1*rand, 1.5 + 0.1*rand) --
      (5 - 1.1 + 0.1*rand, 1.6 + 0.1*rand) --
      (5 - 0.9 + 0.1*rand, 1.5 + 0.1*rand) --
      (5 - 0.6 + 0.1*rand, 1.3 + 0.1*rand) --
      (5 - 0.4 + 0.1*rand, 1.1 + 0.1*rand) --
      (5 - 0.2 + 0.1*rand, 1.4 + 0.1*rand) --
      (5 - 0.5 + 0.1*rand, 1.7 + 0.1*rand) --
      (5 - 0.9 + 0.1*rand, 2.0 + 0.1*rand) --
      (5 - 1.3 + 0.1*rand, 2.2 + 0.1*rand) --
      (5 - 1.2 + 0.1*rand, 3.0 + 0.1*rand) --
      (5 - 0.0 + 0.1*rand, 3.2 + 0.1*rand) --
      (5 - -0.5 + 0.1*rand, 2.6 + 0.1*rand) --
      (5 - -0.7 + 0.1*rand, 1.3 + 0.1*rand) --
      (5 - -0.3 + 0.1*rand, 0.3 + 0.1*rand) --
      cycle;
  \draw [blue,rounded corners=1mm]
      (3.0 + 0.1*rand, 0 + 0.1*rand) --
      (3.2 + 0.1*rand, 1.3 + 0.1*rand) --
      (3.6 + 0.1*rand, 2.7 + 0.1*rand) --
      (3.1 + 0.1*rand, 3.1 + 0.1*rand) --
      (2.6 + 0.1*rand, 3.2 + 0.1*rand) --
      (1.7 + 0.1*rand, 2.6 + 0.1*rand) --
      (1.3 + 0.1*rand, 0.6 + 0.1*rand) --
      (2.0 + 0.1*rand, 0.1 + 0.1*rand) --
      cycle;
  \draw [style=dashed,rounded corners=1mm]
      (3.0, 3.5) --
      (3.7, 2.9) --
      (3.6, 2.2) --
      (3.4, 1.0) --
      (4.2, 0.0);
  \node [] at (3,1.5) { - };
  \node [] at (5,2) { + };

  \node [draw, circle, inner sep=1pt, fill] at (2.8, 2.4) { };
  \node [] at (2.55, 2.4) { $x$ };

  \node [draw, circle, inner sep=1pt, fill] at (4.4, 1.5) { };
  \node [] at (4.65, 1.45) { $x^*$ };

  \draw [style=dotted] (2.8, 2.4) -- (4.4, 1.5) { };
\end{tikzpicture}
\end{center}
\caption{One submanifold has a `pocket'.}
\label{fig:sit2}
\end{subfigure}
\begin{subfigure}[b]{0.3\textwidth}
\begin{center}
\begin{tikzpicture}[scale=0.8]
  \draw [blue,rounded corners=1mm] (2,4) ellipse (2.5cm and 0.7cm);
  \draw [red,rounded corners=1mm] (2,2) ellipse (2.5cm and 0.7cm);
  \draw [style=dashed] (-1.0, 3.1) -- (5.0, 3.1);
  \node [] at (2,4) { - };
  \node [] at (2,2) { + };

  \node [draw, circle, inner sep=1pt, fill] at (2.8, 2.4) { };
  \node [] at (2.55, 2.4) { $x$ };

  \node [draw, circle, inner sep=1pt, fill] at (3.5, 3.3) { };
  \node [] at (3.8, 3.35) { $x^*$ };

  \draw [style=dotted] (2.8, 2.4) -- (3.5, 3.3) { };
\end{tikzpicture}
\end{center}
\caption{Nearby 2D submanifolds.}
\label{fig:sit3}
\end{subfigure}
\end{center}
\caption{{\it (a)}: The adversarial sample $x^*$ is generated by moving off the
`\texttt{-}' submanifold and across the decision boundary (black dashed line), but
$x^*$ still lies far from the `\texttt{+}' submanifold. {\it (b)}: the
`\texttt{+}' submanifold has a `pocket', as in \citet{Szegedy2014}.  $x^*$ lies
in the pocket, presenting significant difficulty for detection. {\it (c)}: the
adversarial sample $x^*$ is near both the decision boundary and both
submanifolds.}
\label{fig:sit}
\end{figure*}

Each of these adversarial sample generation algorithms are able to change the
predicted label of a point without changing the underlying true label: humans
will still correctly classify an adversarial sample, but models will not. This
can be understood from the perspective of the manifold of training data. Many
high-dimensional datasets, such as images, are believed to lie on a low-dimensional
manifold \cite{Lee2007}.
\citet{gardner2015deep} recently showed that by carefully traversing the data
manifold, one can change the underlying true label of an image. 
The intuition is that adversarial perturbations---which do not constitute \textit{meaningful}
changes to the input---must push samples off of the data manifold. 
\citet{Tanay2016} base their investigation of adversarial samples on
the assumption that adversarial samples lie near class boundaries that are close
to the edge of a data submanifold.  Similarly, \citet{Goodfellow2015}
demonstrate that DNNs perform correctly only near the small
manifold of training data.  Therefore, we base our work here on the assumption
that adversarial samples do not lie on the data manifold.

If we accept that adversarial samples are points that would not arise
naturally, then we can assume that a technique to generate adversarial samples
will, from a source point $x$ with class $c_x$, typically generate an
adversarial sample $x^*$ that does not lie on the manifold and is classified
incorrectly as $c_{x^*}$. If $x^*$ lies off of the data manifold, we may split 
into three possible situations:

\begin{enumerate}
  \item $x^*$ is far away from the submanifold of $c_{x^*}$.
  \item $x^*$ is near the submanifold $c_{x^*}$ but not on it, and $x^*$ is far
from the classification boundary separating classes $c_x$ and $c_{x^*}$.
  \item $x^*$ is near the submanifold $c_{x^*}$ but not on it, and $x^*$ is near
the classification boundary separating classes $c_x$ and $c_{x^*}$.
\end{enumerate}

Figures \ref{fig:sit1} through \ref{fig:sit3} show simplified example
illustrations for each of these three situations in a two-dimensional binary
classification setting.

\subsection{Density Estimation}
\label{sec:kde}

If we have an estimate of what the submanifold corresponding to data with class
$c_{x^*}$ is, then we can determine whether $x^*$ falls near this submanifold after
observing the prediction $c_{x^*}$.
Following the intuition of \citet{gardner2015deep} and hypotheses of
\citet{bengio2013better}, the deeper layers of a DNN provide
more linear and `unwrapped' manifolds to work with than input space;
therefore, we may use this idea to model the submanifolds of each class by
performing kernel density estimation in the feature space of the last hidden
layer.

The standard technique of kernel density estimation can, given the point $x$
and the set $X_t$ of training points with label $t$, provide a density
estimate $\hat{f}(x)$ that can be used as a measure of how far $x$ is from
the submanifold for $t$. Specifically,

\vspace*{-1em}
\begin{equation}
\hat{f}(x) = \frac{1}{|X_t|} \sum_{x_i \in X_t} k(x_i, x)
\end{equation}
\vspace*{-1em}

\noindent where $k(\cdot, \cdot)$ is the kernel function, often chosen as a
Gaussian with bandwidth $\sigma$:

\vspace*{-1em}
\begin{equation}
k_\sigma(x, y) \sim \exp(-\| x - y \|^2 / \sigma^2).
\end{equation}
\vspace*{-1.5em}

\begin{figure}[b!]
\begin{center}
  \pgfmathdeclarefunction{gauss}{2}{%
  \pgfmathparse{1/(#2*sqrt(2*pi))*exp(-((x-#1)^2)/(2*#2^2))}%
}

\begin{tikzpicture}[scale=1.0]
  \begin{axis}[every axis plot post/.append style={
      mark=none,domain=-2:4,samples=50,smooth},
      x tick label style={major tick length=0pt},
      axis x line=middle,
      axis y line=none,
      axis line style={<->},
      xticklabels={},
      ymin=-0.5,
      enlargelimits=upper,
      width=0.48\textwidth,
      height=0.2\textwidth]
  \addplot{gauss(-0.3, 0.6) +
           gauss(0.3, 0.6) +
           gauss(2.1, 0.6) +
           gauss(2.35, 0.6) +
           gauss(2.7, 0.6)};
  \node[circle,fill,inner sep=1pt] at (axis cs:-0.3,0) {};
  \node[circle,fill,inner sep=1pt] at (axis cs:0.3,0) {};
  \node[circle,fill,inner sep=1pt] at (axis cs:2.1,0) {};
  \node[circle,fill,inner sep=1pt] at (axis cs:2.35,0) {};
  \node[circle,fill,inner sep=1pt] at (axis cs:2.7,0) {};
\end{axis}
\end{tikzpicture}
\end{center}
\vspace*{-1.7em}
\caption{Overly smooth density estimate from a too-large bandwidth on 1-D points
sampled from a bimodal distribution.}
\vspace*{-1em}
\label{fig:smooth}
\end{figure}
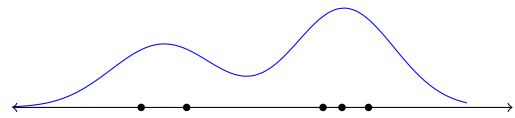

The bandwidth may typically be chosen as a value that maximizes the
log-likelihood of the training data \citep{jones1996brief}. A value too small
will give rise to a `spiky' density estimate with too many gaps (see
Figure \ref{fig:spiky}), but a value too large will give rise to an
overly-smooth density estimate (see Figure \ref{fig:smooth}). This also
implies that the estimate is improved as the training set size
$|X_t|$ increases, since we are able to use smaller bandwidths without the
estimate becoming too `spiky.'

\begin{figure}[b!]
\begin{center}
  \pgfmathdeclarefunction{gauss}{2}{%
  \pgfmathparse{1/(#2*sqrt(2*pi))*exp(-((x-#1)^2)/(2*#2^2))}%
}

\begin{tikzpicture}[scale=1.0]
  \begin{axis}[every axis plot post/.append style={
      mark=none,domain=-2:4,samples=50,smooth},
      x tick label style={major tick length=0pt},
      axis x line=middle,
      axis y line=none,
      axis line style={<->},
      xticklabels={},
      enlargelimits=upper,
      ymin=-0.5,
      width=0.48\textwidth,
      height=0.2\textwidth]
  \addplot{gauss(-0.3, 0.1) +
           gauss(0.3, 0.1) +
           gauss(2.1, 0.1) +
           gauss(2.35, 0.1) +
           gauss(2.7, 0.1)};
  \node[circle,fill,inner sep=1pt] at (axis cs:-0.3,0) {};
  \node[circle,fill,inner sep=1pt] at (axis cs:0.3,0) {};
  \node[circle,fill,inner sep=1pt] at (axis cs:2.1,0) {};
  \node[circle,fill,inner sep=1pt] at (axis cs:2.35,0) {};
  \node[circle,fill,inner sep=1pt] at (axis cs:2.7,0) {};
\end{axis}
\end{tikzpicture}
\end{center}
\vspace*{-1.5em}
\caption{`spiky' density estimate from a too-small bandwidth on 1-D points sampled
from a bimodal distribution.}
\vspace*{-0.6em}
\label{fig:spiky}
\end{figure}
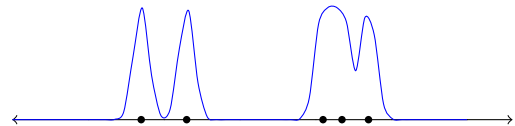

For the manifold estimate, we operate in the space of the last hidden layer. 
This layer provides a space of reasonable dimensionality in which we expect the
manifold of our data to be simplified.
If $\phi(x)$ is
the last hidden layer activation vector for point $x$, then our density estimate
for a point $x$ with predicted class $t$ is defined as

\vspace*{-1em}
\begin{equation}
\hat{K}(x, X_t) = \sum_{x_i \in X_t} k_\sigma(\phi(x), \phi(x_i))
\end{equation}
\vspace*{-1em}

\noindent where $X_t$ is the set of training points of class $t$, and $\sigma$ is the tuned bandwidth.

\begin{figure}[t!]
\begin{subfigure}[b]{0.235\textwidth}
\begin{center}
\begin{tikzpicture}[scale=1.0]
  \begin{axis}[scale only axis,
      axis x line*=bottom,
      axis y line*=left,
      axis line style={->},
      ymin=-180,
      ymax=-150,
      xmin=1,
      xmax=50,
      yticklabels={},
      xticklabels={},
      x label style={at={(axis description cs:0.5,0.24)},anchor=north},
      y label style={at={(axis description cs:0.34,0.5)},anchor=south east,rotate=-90},
      xlabel={\footnotesize{iteration}},
      ylabel={\footnotesize{$\hat{K}$}},
      enlargelimits=upper,
      width=0.9\textwidth,
      height=0.4\textwidth]

\addplot[color=blue] coordinates {
(1, -1.520308368949386875e+02)
(2, -1.523591027255152994e+02)
(3, -1.527329625028610280e+02)
(4, -1.531035745944065525e+02)
(5, -1.533903617951957585e+02)
(6, -1.536495831009305277e+02)
(7, -1.538687364603734125e+02)
(8, -1.540849070146457791e+02)
(9, -1.543232346800365917e+02)
(10, -1.545668577683692888e+02)
(11, -1.547783647488100200e+02)
(12, -1.549540366240426579e+02)
(13, -1.551192640760142751e+02)
(14, -1.552361921179143565e+02)
(15, -1.553986179010571504e+02)
(16, -1.555955202364225158e+02)
(17, -1.558357487255761669e+02)
(18, -1.562673047559801773e+02)
(19, -1.568074292075095570e+02)
(20, -1.574659613675242724e+02)
(21, -1.581657988878578749e+02)
(22, -1.589679896700024528e+02)
(23, -1.598827353083818821e+02)
(24, -1.608594957174594242e+02)
(25, -1.619267033804085543e+02)
(26, -1.628112031378253732e+02)
(27, -1.641123490161789675e+02)
(28, -1.653788462083032584e+02)
(29, -1.668523965437334766e+02)
(30, -1.683759726119276365e+02)
(31, -1.691728552638447525e+02)
(32, -1.697575615144391179e+02)
(33, -1.702951355787856471e+02)
(34, -1.707357471595272216e+02)
(35, -1.711724408683959950e+02)
(36, -1.716506588164288019e+02)
(37, -1.719125599918937155e+02)
(38, -1.722461262745593160e+02)
(39, -1.726353308356672756e+02)
(40, -1.729721821218847140e+02)
(41, -1.733671027120730344e+02)
(42, -1.736696475749538706e+02)
(43, -1.740698470514208793e+02)
(44, -1.743558975899685208e+02)
(45, -1.748200024851711305e+02)
(46, -1.749593624362206015e+02)
(47, -1.753183964703390529e+02)
(48, -1.755071389272471265e+02)
(49, -1.757215503258633191e+02)
(50, -1.758759304795227081e+02)
};

\addplot[color=red] coordinates {
(1, -1.718411563166096983e+02)
(2, -1.705311894269547679e+02)
(3, -1.692974849655276159e+02)
(4, -1.682236622389906699e+02)
(5, -1.671917074923522932e+02)
(6, -1.663077887709852689e+02)
(7, -1.654581258093395775e+02)
(8, -1.647720548554384550e+02)
(9, -1.641157665568469213e+02)
(10, -1.634704966593956499e+02)
(11, -1.628435053594555484e+02)
(12, -1.621774001466043273e+02)
(13, -1.614957755674923590e+02)
(14, -1.608284279653566955e+02)
(15, -1.602196915019857784e+02)
(16, -1.596555518587198037e+02)
(17, -1.590771800909799936e+02)
(18, -1.586483905230832363e+02)
(19, -1.583691932973623864e+02)
(20, -1.582027563745525356e+02)
(21, -1.581362610108059812e+02)
(22, -1.581582248763063774e+02)
(23, -1.582563210373001539e+02)
(24, -1.584572947638984317e+02)
(25, -1.587506986701799008e+02)
(26, -1.589136904424219949e+02)
(27, -1.591001882608360063e+02)
(28, -1.592666511744192519e+02)
(29, -1.592292707351914487e+02)
(30, -1.588530802445939969e+02)
(31, -1.586846402376566232e+02)
(32, -1.584334052754619222e+02)
(33, -1.581833460062032373e+02)
(34, -1.580539815897971039e+02)
(35, -1.579201643537847133e+02)
(36, -1.577551455544932537e+02)
(37, -1.575848639179250199e+02)
(38, -1.574511709694072294e+02)
(39, -1.573733299256553835e+02)
(40, -1.572585756946184858e+02)
(41, -1.572289636014069174e+02)
(42, -1.571373431532650216e+02)
(43, -1.570980409875535315e+02)
(44, -1.570578777014771674e+02)
(45, -1.570589530939348890e+02)
(46, -1.570062046452887046e+02)
(47, -1.569976154608825709e+02)
(48, -1.570218534463206481e+02)
(49, -1.569407103801564176e+02)
(50, -1.569257657253590708e+02)
};

\addplot[color=black] coordinates {
(19, -1000)
(19, 1000)
};
\end{axis}
\end{tikzpicture}
\end{center}
\end{subfigure}
\begin{subfigure}[b]{0.235\textwidth}
\begin{center}
\begin{tikzpicture}[scale=1.0]
  \begin{axis}[scale only axis,
      axis x line*=bottom,
      axis y line*=left,
      axis line style={->},
      ymin=-180,
      ymax=-150,
      xmin=1,
      xmax=50,
      yticklabels={},
      xticklabels={},
      x label style={at={(axis description cs:0.5,0.24)},anchor=north},
      y label style={at={(axis description cs:0.34,0.5)},anchor=south east,rotate=-90},
      xlabel={\footnotesize{iteration}},
      ylabel={\footnotesize{$\hat{K}$}},
      enlargelimits=upper,
      width=0.9\textwidth,
      height=0.4\textwidth]

\addplot[color=blue] coordinates {
(1, -1.497554745482801195e+02)
(2, -1.499617991564859665e+02)
(3, -1.502357459419612553e+02)
(4, -1.505856207388296752e+02)
(5, -1.509506936948587850e+02)
(6, -1.513697874299536466e+02)
(7, -1.518294203599514560e+02)
(8, -1.524160271873195995e+02)
(9, -1.531433214507607374e+02)
(10, -1.539278952585641491e+02)
(11, -1.546980975999280759e+02)
(12, -1.554545931284667120e+02)
(13, -1.561400744759730230e+02)
(14, -1.567779630704844465e+02)
(15, -1.574657607218760234e+02)
(16, -1.582024533136278990e+02)
(17, -1.589047235375291507e+02)
(18, -1.595524739184282055e+02)
(19, -1.601297977243675632e+02)
(20, -1.606647921431511747e+02)
(21, -1.612434895743833465e+02)
(22, -1.618667701944261808e+02)
(23, -1.624547445266568104e+02)
(24, -1.631087347162161052e+02)
(25, -1.633852608743441692e+02)
(26, -1.635533590518848257e+02)
(27, -1.636637537953920400e+02)
(28, -1.639405619753536882e+02)
(29, -1.643422205835776708e+02)
(30, -1.648716964323895979e+02)
(31, -1.651514643754169356e+02)
(32, -1.653550070724191130e+02)
(33, -1.655805202343383939e+02)
(34, -1.657981952600074180e+02)
(35, -1.660485281080865150e+02)
(36, -1.662240716125379549e+02)
(37, -1.665321978560924663e+02)
(38, -1.665961442073069065e+02)
(39, -1.668614452781514501e+02)
(40, -1.669474189858339059e+02)
(41, -1.671891977409247829e+02)
(42, -1.672843880217549781e+02)
(43, -1.676847518940746227e+02)
(44, -1.675835309982723231e+02)
(45, -1.679134753127545991e+02)
(46, -1.679303970791707741e+02)
(47, -1.681504705527535748e+02)
(48, -1.681303510021660088e+02)
(49, -1.683749153595566099e+02)
(50, -1.682132225527208789e+02)
};

\addplot[color=red] coordinates {
(1, -1.603585171981488884e+02)
(2, -1.598827090572335976e+02)
(3, -1.593120368476054978e+02)
(4, -1.586740812696180853e+02)
(5, -1.581339676738829780e+02)
(6, -1.577785460086631133e+02)
(7, -1.576091221295421008e+02)
(8, -1.576801057463096072e+02)
(9, -1.579626091457602968e+02)
(10, -1.583761419202404852e+02)
(11, -1.586754625063666708e+02)
(12, -1.583968055942004298e+02)
(13, -1.576267161809522008e+02)
(14, -1.568526372877626045e+02)
(15, -1.561373176452079008e+02)
(16, -1.556611731133975809e+02)
(17, -1.553855579300675629e+02)
(18, -1.553810210646737744e+02)
(19, -1.555155175224293203e+02)
(20, -1.558227491721617923e+02)
(21, -1.563081994106397019e+02)
(22, -1.568735709529668156e+02)
(23, -1.572329449323756876e+02)
(24, -1.574685858733128612e+02)
(25, -1.573829785844480398e+02)
(26, -1.571563914825709958e+02)
(27, -1.568596716035763734e+02)
(28, -1.565260776832282374e+02)
(29, -1.563648691494791763e+02)
(30, -1.562743857668943690e+02)
(31, -1.561539258987229744e+02)
(32, -1.560894215845482336e+02)
(33, -1.559664194164542437e+02)
(34, -1.559083459907028555e+02)
(35, -1.557801829825174593e+02)
(36, -1.556558612599155254e+02)
(37, -1.556504371560582456e+02)
(38, -1.555171865715283275e+02)
(39, -1.553947890617936309e+02)
(40, -1.552737164802443601e+02)
(41, -1.551871297691177745e+02)
(42, -1.551073250857585606e+02)
(43, -1.551348772768303377e+02)
(44, -1.550119950974597884e+02)
(45, -1.550949732169914910e+02)
(46, -1.549947441456669708e+02)
(47, -1.549986607298698118e+02)
(48, -1.549434452877474371e+02)
(49, -1.550210248422214647e+02)
(50, -1.549508472880071395e+02)
};

\addplot[color=black] coordinates {
(16, -1000)
(16, 1000)
};
\end{axis}
\end{tikzpicture}
\end{center}
\end{subfigure}
\begin{subfigure}[b]{0.235\textwidth}
\begin{center}
\begin{tikzpicture}[scale=1.0]
  \begin{axis}[scale only axis,
      axis x line*=bottom,
      axis y line*=left,
      axis line style={->},
      ymin=-180,
      ymax=-150,
      xmin=1,
      xmax=50,
      xticklabels={},
      yticklabels={},
      x label style={at={(axis description cs:0.5,0.24)},anchor=north},
      y label style={at={(axis description cs:0.34,0.5)},anchor=south east,rotate=-90},
      xlabel={\footnotesize{iteration}},
      ylabel={\footnotesize{$\hat{K}$}},
      enlargelimits=upper,
      width=0.9\textwidth,
      height=0.4\textwidth]

\addplot[color=blue] coordinates {
(1, -1.509569840047603577e+02)
(2, -1.507136310487313722e+02)
(3, -1.506081260305954004e+02)
(4, -1.506298552631026553e+02)
(5, -1.507549136564387595e+02)
(6, -1.508989634689567652e+02)
(7, -1.510468617222942669e+02)
(8, -1.512659694856755550e+02)
(9, -1.516185313227026654e+02)
(10, -1.521383007246138845e+02)
(11, -1.527519673475075592e+02)
(12, -1.534263918528038459e+02)
(13, -1.539566095136633237e+02)
(14, -1.543166223897750911e+02)
(15, -1.547301134440686781e+02)
(16, -1.550368220196315008e+02)
(17, -1.553842600143284756e+02)
(18, -1.558005964482995296e+02)
(19, -1.562073890328356072e+02)
(20, -1.567390070156856723e+02)
(21, -1.574796075563313877e+02)
(22, -1.582321759421639626e+02)
(23, -1.589906814036943388e+02)
(24, -1.597231970196501152e+02)
(25, -1.604438593357884031e+02)
(26, -1.609720325531829417e+02)
(27, -1.614478255298194540e+02)
(28, -1.616922027899041723e+02)
(29, -1.619961130062225436e+02)
(30, -1.622420533595529548e+02)
(31, -1.619836959419848483e+02)
(32, -1.621036182990401073e+02)
(33, -1.619545249509473592e+02)
(34, -1.619290869891624141e+02)
(35, -1.618199416435853379e+02)
(36, -1.619196443745428269e+02)
(37, -1.616634894144987982e+02)
(38, -1.619086523086164959e+02)
(39, -1.617171971883791173e+02)
(40, -1.620181947324135763e+02)
(41, -1.618916496419319344e+02)
(42, -1.621938751300995705e+02)
(43, -1.619148070691125270e+02)
(44, -1.622432985340909397e+02)
(45, -1.621575192755179842e+02)
(46, -1.623499391509927534e+02)
(47, -1.621786115965910255e+02)
(48, -1.623513921485375135e+02)
(49, -1.624844367446882814e+02)
(50, -1.625133287001014253e+02)
};

\addplot[color=red] coordinates {
(1, -1.698664052463311975e+02)
(2, -1.678439456101702945e+02)
(3, -1.660563812911473178e+02)
(4, -1.645061679767353837e+02)
(5, -1.631585909708502413e+02)
(6, -1.621245810649105294e+02)
(7, -1.614436157567734256e+02)
(8, -1.611247714197875496e+02)
(9, -1.613542711199979749e+02)
(10, -1.619013018356671694e+02)
(11, -1.624966417130331138e+02)
(12, -1.620976975949998007e+02)
(13, -1.609970440479612535e+02)
(14, -1.597307470867777681e+02)
(15, -1.583341043969741690e+02)
(16, -1.569428443452837882e+02)
(17, -1.558378379353423213e+02)
(18, -1.549813287888563025e+02)
(19, -1.543346963664717748e+02)
(20, -1.538095074151966060e+02)
(21, -1.533984147008419257e+02)
(22, -1.532160500213124976e+02)
(23, -1.530252345618746403e+02)
(24, -1.529079865689727740e+02)
(25, -1.527546473839900045e+02)
(26, -1.525169140313776950e+02)
(27, -1.524227777964294717e+02)
(28, -1.522985291917652546e+02)
(29, -1.522571187457515975e+02)
(30, -1.522821277158052737e+02)
(31, -1.520661604922208312e+02)
(32, -1.521239524099448772e+02)
(33, -1.519825904054085015e+02)
(34, -1.519954688133225318e+02)
(35, -1.519256968297502510e+02)
(36, -1.520217616468221422e+02)
(37, -1.518506442284062814e+02)
(38, -1.520366059261626788e+02)
(39, -1.518941908193166910e+02)
(40, -1.520760184654209866e+02)
(41, -1.519703970562337361e+02)
(42, -1.521804683531223361e+02)
(43, -1.520043914814362438e+02)
(44, -1.521854009028779160e+02)
(45, -1.521144322732015439e+02)
(46, -1.522399578262646287e+02)
(47, -1.521011467854432624e+02)
(48, -1.522299375097636300e+02)
(49, -1.522932033068473174e+02)
(50, -1.523319740135146958e+02)
};

\addplot[color=black] coordinates {
(16, -1000)
(16, 1000)
};
\end{axis}
\end{tikzpicture}
\end{center}
\end{subfigure}
\begin{subfigure}[b]{0.235\textwidth}
\begin{center}
\begin{tikzpicture}[scale=1.0]
  \begin{axis}[scale only axis,
      axis x line*=bottom,
      axis y line*=left,
      axis line style={->},
      ymin=-180,
      ymax=-150,
      xmin=1,
      xmax=50,
      xticklabels={},
      yticklabels={},
      x label style={at={(axis description cs:0.5,0.24)},anchor=north},
      y label style={at={(axis description cs:0.34,0.5)},anchor=south east,rotate=-90},
      xlabel={\footnotesize{iteration}},
      ylabel={\footnotesize{$\hat{K}$}},
      enlargelimits=upper,
      width=0.9\textwidth,
      height=0.4\textwidth]

\addplot[color=blue] coordinates {
(1, -1.547238760830583431e+02)
(2, -1.551101975376957967e+02)
(3, -1.555496867640049174e+02)
(4, -1.560247781249783827e+02)
(5, -1.564903362497904880e+02)
(6, -1.569543101654037969e+02)
(7, -1.574436526845285016e+02)
(8, -1.579178035545870671e+02)
(9, -1.583028445827324049e+02)
(10, -1.585648974558989437e+02)
(11, -1.588105710568201800e+02)
(12, -1.592073695264531921e+02)
(13, -1.598594171225616662e+02)
(14, -1.607139826173488473e+02)
(15, -1.616428581370714710e+02)
(16, -1.625900661344958280e+02)
(17, -1.637124058325003659e+02)
(18, -1.650317318038082419e+02)
(19, -1.664353171965456966e+02)
(20, -1.679834485733744316e+02)
(21, -1.694380657278065598e+02)
(22, -1.708454785600588082e+02)
(23, -1.723995043900072233e+02)
(24, -1.740712499256316903e+02)
(25, -1.758861472484153126e+02)
(26, -1.776102642112555543e+02)
(27, -1.785940316073936174e+02)
(28, -1.815211193854837575e+02)
(29, -1.796550181641255790e+02)
(30, -1.801718422509368622e+02)
(31, -1.863078481309096333e+02)
(32, -1.869748097489184602e+02)
(33, -1.816883906771069803e+02)
(34, -1.880774984816167432e+02)
(35, -1.813175291822047370e+02)
(36, -1.892462345051138470e+02)
(37, -1.818313704757837570e+02)
(38, -1.904061110920420106e+02)
(39, -1.838069761287143820e+02)
(40, -1.913563428018764512e+02)
(41, -1.917846795315321629e+02)
(42, -1.833359095822554821e+02)
(43, -1.856398598576061545e+02)
(44, -1.823974878872095644e+02)
(45, -1.845101428558911891e+02)
(46, -1.828874377259529638e+02)
(47, -1.935173383000688148e+02)
(48, -1.937108507677271234e+02)
(49, -1.827499182453862829e+02)
(50, -1.940593687868883137e+02)
};

\addplot[color=red] coordinates {
(1, -1.792719993901332032e+02)
(2, -1.918543761814029835e+02)
(3, -1.900744668324344957e+02)
(4, -1.882913717141251482e+02)
(5, -1.863717098664544380e+02)
(6, -1.786926525722096244e+02)
(7, -1.777193338265956015e+02)
(8, -1.769091759313072885e+02)
(9, -1.763213538987431832e+02)
(10, -1.752348103680446627e+02)
(11, -1.735982223087010254e+02)
(12, -1.718275183240859292e+02)
(13, -1.699757492484871193e+02)
(14, -1.683687467429749063e+02)
(15, -1.669791873583946256e+02)
(16, -1.656583917192505737e+02)
(17, -1.644092488449165899e+02)
(18, -1.631625941405210654e+02)
(19, -1.621099577489491423e+02)
(20, -1.612152139855667770e+02)
(21, -1.606233110546116905e+02)
(22, -1.602075592862926783e+02)
(23, -1.600216116815614384e+02)
(24, -1.599788982113573468e+02)
(25, -1.601858987255048703e+02)
(26, -1.604810897394301890e+02)
(27, -1.607440251793241259e+02)
(28, -1.608319512413959274e+02)
(29, -1.606997659002826424e+02)
(30, -1.604566262426044432e+02)
(31, -1.604314259549076951e+02)
(32, -1.604034219013960865e+02)
(33, -1.604048348845844316e+02)
(34, -1.603795997253742769e+02)
(35, -1.603966774785250209e+02)
(36, -1.603831315236570561e+02)
(37, -1.604220230063080237e+02)
(38, -1.603969046288599145e+02)
(39, -1.604490403407759800e+02)
(40, -1.604719182135916924e+02)
(41, -1.604681693486483312e+02)
(42, -1.604420184481083425e+02)
(43, -1.604673857513367068e+02)
(44, -1.604100343711873222e+02)
(45, -1.604584293049893517e+02)
(46, -1.604307802991608867e+02)
(47, -1.604031241651142636e+02)
(48, -1.603933619028349540e+02)
(49, -1.603833763083093231e+02)
(50, -1.603680314069237909e+02)
};

\addplot[color=black] coordinates {
(16, -1000)
(16, 1000)
};
\end{axis}
\end{tikzpicture}
\end{center}
\end{subfigure}
\vspace*{-0.7em}
\caption{Density estimate as a function of number of iterations of the BIM
attack for a few MNIST sample points.  The estimate decreases for the source
class (blue) and increases for the incorrect class (red); usually the crossover
point is near when the predicted class changes (black line).}
\label{fig:kde_walk}
\vspace*{-0.5em}
\end{figure}
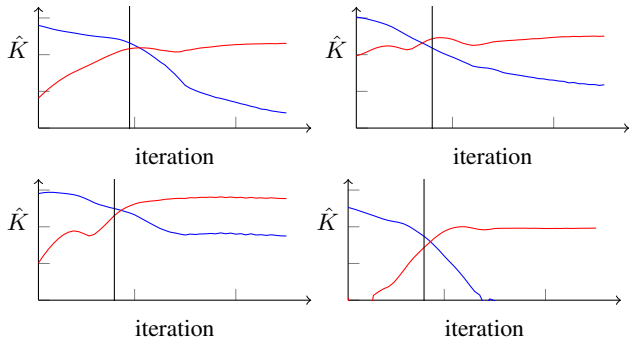

To validate our intuition about the utility of this density estimate, we perform
a toy experiment using the BIM attack with a convnet trained on MNIST data.
In Figure \ref{fig:kde_walk}, we plot the density estimate of the source class
and the final predicted class for each iteration of BIM. One can see that the
adversarial sample moves away from a high density estimate region for the
correct class, and towards a high density estimate region for the incorrect
class. This matches our intuition: we expect the adversarial sample to leave
the correct class manifold and move towards (but not onto) the incorrect class
manifold.

While a density estimation approach can easily detect an adversarial
point that is far from the $c_{x^*}$ submanifold, this strategy may not work well
when $x^*$ is very near the $c_{x^*}$ submanifold. Therefore, we must investigate
alternative approaches for those cases.

\subsection{Bayesian Neural Network Uncertainty}
\label{sec:bayesian_uncert}

\begin{figure}[t!]
    \includegraphics[width=0.45\textwidth]{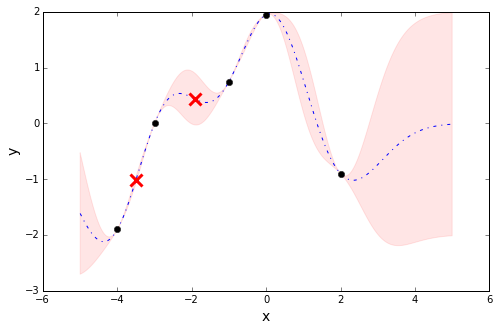}
    \vspace*{-1.5em}
    \caption{A simple 1-dimensional Gaussian process regression. The dashed line
indicates the mean prediction and the shaded area indicates the 95\% confidence
interval. While two test points (red x's) are equidistant from their nearest training
points (black dots), their uncertainty estimates differ significantly.}
    \label{fig:gaussian_process}
\end{figure}

Beyond distance-based metrics, another powerful
tool to identify low-confidence regions of the input space is 
the uncertainty output of Bayesian models, e.g., the Gaussian process 
\cite{Rasmussen2005}. Gaussian processes assume a Gaussian prior over the set of 
all functions, $\mathcal{F}$, that can be used to map the input space to the 
output space. As observations ($x,y$) are made, only those functions 
$f\in\mathcal{F}$ are retained for which $f(x)=y$. For a new test point $x^*$, the 
prediction for each function $f$, $y^*=f(x^*)$, is computed and the expected value 
over $y^*$ is the used as the final prediction. Simultaneously, the variance of the 
output values $y^*$ is also used as an indicator of the model's uncertainty. Figure 
\ref{fig:gaussian_process} illustrates how in simple cases, Bayesian uncertainty can 
provide additional information about model confidence not conveyed by distance 
metrics like a density estimate.

Recently, \citet{Gal2015} proved that DNNs trained with dropout are equivalent to an approximation
of the deep Gaussian process. As result, we can extract Bayesian uncertainty estimates
from a wide range of DNN architectures without modification. Dropout, first introduced as a method to
reduce overfitting when training DNNs \cite{Srivastava2014}, works by dropping hidden nodes from the network 
randomly with some probability $p$ during the training phase. During the testing phase, 
all nodes are kept, but the weights are scaled by $p$. \citet{Gal2015} showed that the dropout training objective 
converges to a minimization of the Kullback-Leibler divergence between an aproximate distribution 
and the posterior of a deep Gaussian process marginalized over its covariance function parameters.
After iterating to convergence, uncertainty estimates can be extracted from dropout DNNs in 
the following manner.

We sample $T$ times from our distribution of network configurations, typically i.i.d. $Bernoulli(o_l)$ 
for each layer $l$, and obtain parameters $\{W^1, \cdots, W^T\}$. Here $W^t = \{W_1^t,...,W_L^t\}$ 
are the $L$ weight matrices sampled at iteration $t$. Thereafter, we can evaluate a Monte Carlo 
estimate of the output, i.e. the first moment, as:

\vspace*{-1.3em}
\begin{equation}
\mathbb{E}_{q(y^*|x^*)}[y^*] \approx \frac{1}{T}\sum\limits_{i=1}^T\hat{y}^*(x^*,W^t).
\label{eq:first_moment}
\end{equation}
\vspace*{-1.3em}

Similarly, we can evaluate the second moment with Monte Carlo estimation, 
leading us to an estimate of model variance

\vspace*{-1.7em}
\begin{eqnarray}
\mathbb{V}_{q(y^*|x^*)}[y^*] &\approx& \tau^{-1}I_D \nonumber \\ 
&&+ \frac{1}{T} \sum\limits_{i=1}^T \hat{y}^*(x^*,W^t)^T
\hat{y}^*(x^*,W^t)\nonumber \\
&& - \mathbb{E}_{q(y^*|x^*)}[y^*]^T \mathbb{E}_{q(y^*|x^*)}[y^*]
\label{eq:variance_approx}
\end{eqnarray}
\vspace*{-1.7em}

where $\tau$ is our model precision. 

Relying on the intuition that Bayesian uncertainty can be useful to identify adversarial samples,
we make use of dropout variance values in this paper, setting $T=50$. As dropout on its own is
known to be a powerful regularization technique \cite{Srivastava2014}, we use 
neural network models without weight decay, leaving $\tau^{-1}=0$. Thus, for a test sample
$x^*$ and stochastic predictions $\{\hat{y}^*_1,...,\hat{y}^*_T\}$, 
our uncertainty estimate $U(x^*)$ can be computed as

\vspace*{-2.0em}
\begin{equation}
U(x^*) =
\frac{1}{T} \sum\limits_{i=1}^{T} \hat{y_i^*}^T \hat{y_i^*}
- \left(\frac{1}{T} \sum\limits_{i=1}^{T} \hat{y_i^*}\right)^T \left(\frac{1}{T} \sum\limits_{i=1}^{T} \hat{y_i^*}\right).
\label{eq:uncertainty}
\end{equation}
\vspace*{-0.3em}

Because we use DNNs with one output node per class, we look at the mean of the uncertainty 
vector as a scalar representation of model uncertainty.

\begin{figure}[t!]
\begin{subfigure}[b]{0.5\textwidth}
\begin{center}
\includegraphics[width=0.98\textwidth]{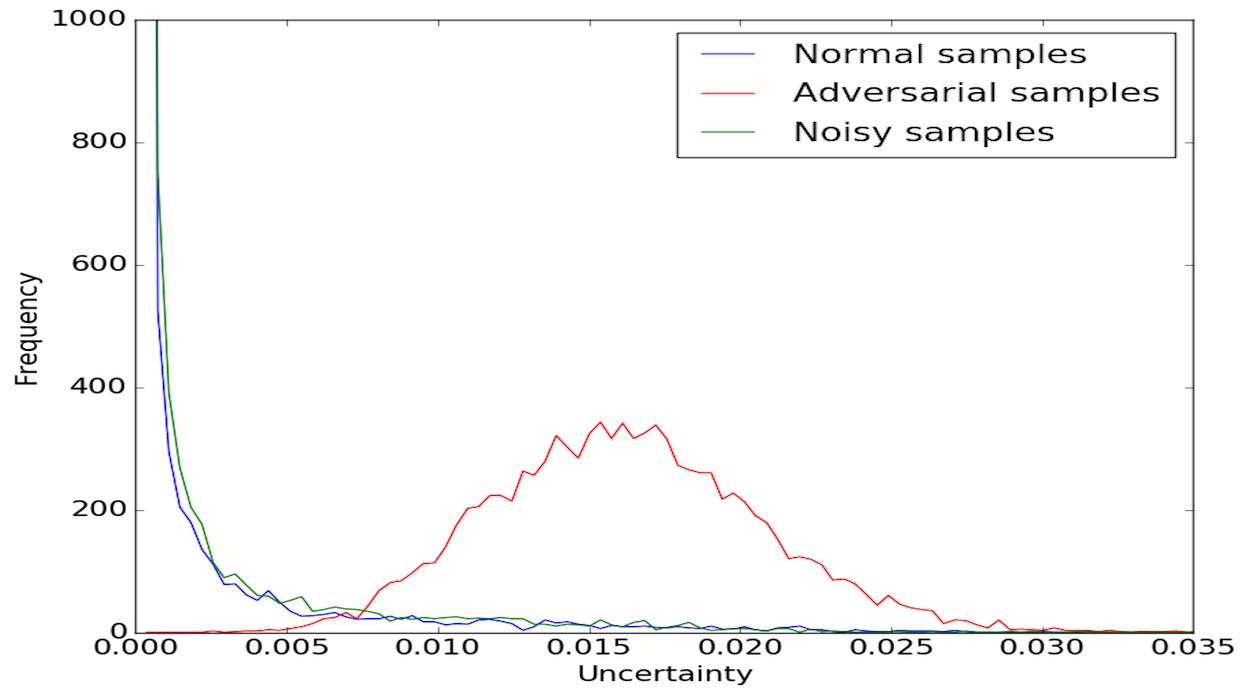}
\end{center}
\vspace*{-0.8em}
\caption{BIM}
\label{fig:hist_bim}
\end{subfigure}
\begin{subfigure}[b]{0.5\textwidth}
\begin{center}
\includegraphics[width=0.98\textwidth]{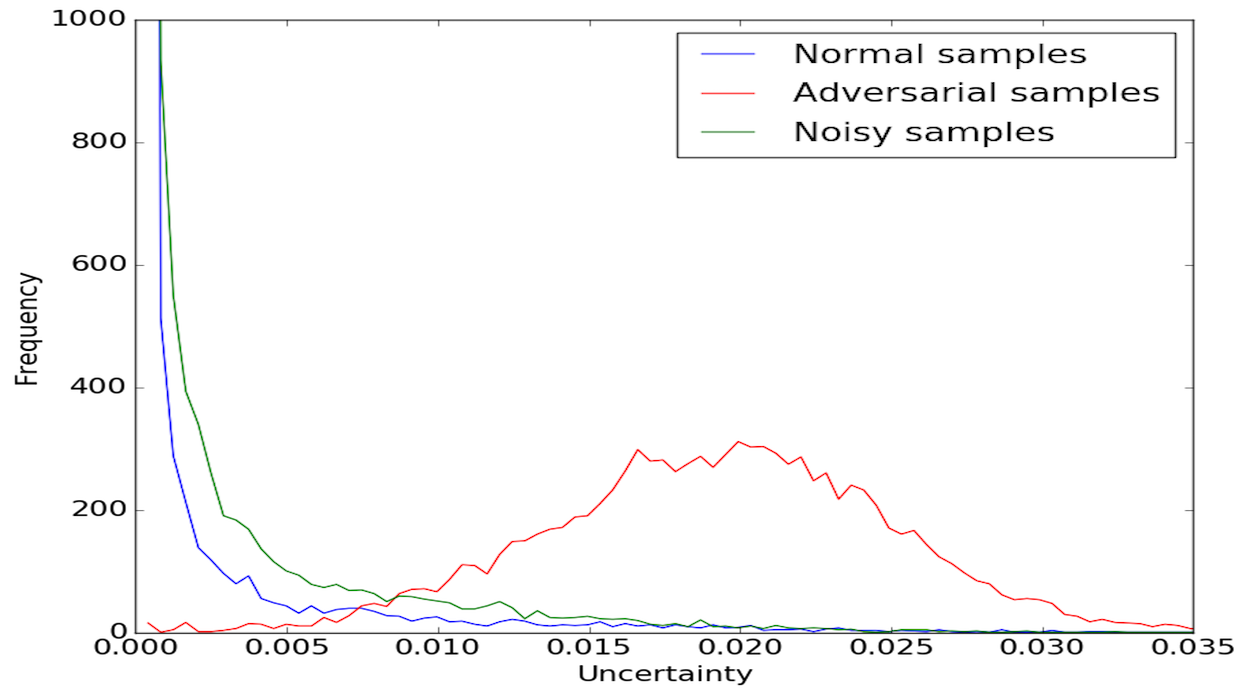}
\end{center}
\vspace*{-0.8em}
\caption{JSMA}
\label{fig:hist_jsma}
\end{subfigure}
\vspace*{-2em}
\caption{Model uncertainty distributions per sample type for MNIST. Distributions are based on a histogram with 100 bins.}
\label{fig:hist}
\vspace*{-0.8em}
\end{figure}

To demonstrate the efficacy of our uncertainty estimates in detecting adversarial samples, we 
trained the LeNet convnet \cite{lecun1989backpropagation} with a dropout rate of 0.5 applied after the last pooling layer and after the
inner-product layer for MNIST classification. Figures \ref{fig:hist_bim} and \ref{fig:hist_jsma}
compare the distribution of Bayesian uncertainty for adversarial samples to those of normal samples and of noisy
samples with equivalent perturbation size; both the BIM and JSMA cases are shown. Clearly, uncertainty distributions for 
adversarial samples are statistically distinct from normal and noisy samples, verifying our intuition.

\vspace*{-0.4em}
\section{Experiments}
\label{sec:experiments}
\vspace*{-0.2em}

\begin{table*}[t!]
\begin{center}
\begin{tabular}{|c|c|c|c|c|c|c|c|c|c|c|}
\hline
{\bf Dataset} & \multicolumn{2}{|c|}{\bf FGSM} & \multicolumn{2}{|c|}{\bf BIM-A}
& \multicolumn{2}{|c|}{\bf BIM-B} & \multicolumn{2}{|c|}{\bf JSMA} &
\multicolumn{2}{|c|}{\bf C\&W} \\
& $L_2$ & Acc. & $L_2$ & Acc. & $L_2$ & Acc. & $L_2$ & Acc. & $L_2$ & Acc. \\
\hline
MNIST & $6.22$ & $5.87$\% & $2.62$ & $0.00$\% & $5.37$ & $0.00$\% & $5.00$ &
$2.70$\% & $4.71$ & $0.79$\% \\
CIFAR-10 & $2.74$ & $7.03$\% & $0.48$ & $0.57$\% & $2.14$ & $0.57$\% & $3.45$ &
$0.20$\% & $2.70$ & $0.89$\% \\
SVHN & $7.08$ & $3.29$\% & $0.83$ & $0.00$\% & $6.56$ & $0.00$\% & $2.96$ &
$0.32$\% & $2.37$ & $0.87$\% \\
\hline
\end{tabular}
\end{center}
\caption{Adversarial attack details. For each algorithm, the average $L_2$-norm of
the perturbation is shown,
as well as the model accuracy on the adversarial set.}
\label{tab:attack_details}
\end{table*}

In order to evaluate the proficiency of our density and uncertainty features for
adversarial detection, we test these features on MNIST, CIFAR10, and SVHN. All pixels are
scaled to floats in the range of [0, 1]. Our models achieve near
state-of-the-art accuracy on the normal holdout sets for each dataset and are
described in Section \ref{sec:network_setup}. In order to properly evaluate our method,
we only perturb those test samples which were correctly classified by our models in
their original states. An adversary would have no reason to perturb samples that are
already misclassified.

\begin{figure}[b!]
\begin{center}
  \includegraphics[width=0.47\textwidth]{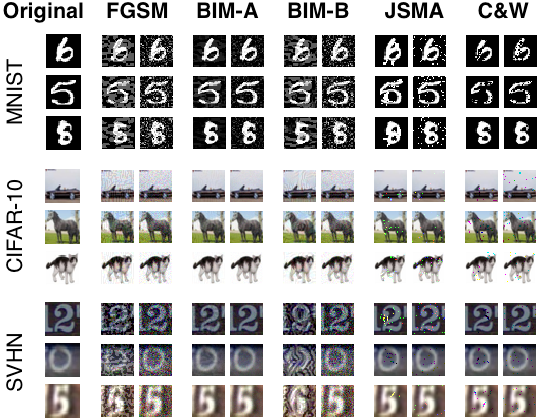}
\end{center}
  \vspace*{-1em}
 \caption{Some example images from MNIST, CIFAR-10 and SVHN. The original image is shown in the left-most column. For each attack, the left image is the adversarial sample and the right image is a corresponding noisy sample of equal perturbation size.}
 \label{fig:normal_noisy_adv}
\end{figure}

\begin{table*}[t!]
\begin{center}
\begin{tabular}{|c|c|c|c|c|c|c|c|c|}
\hline
{\bf Sample } & \multicolumn{4}{|c|}{\bf MNIST} & \multicolumn{4}{|c|}{\bf CIFAR-10} \\
\bf Type & $\frac{u(x^*)}{u(x)} >1 $ &  $\frac{d(x^*)} {d(x)} <1 $ & $\frac{u(x^*)}{u(x^n)} > 1$ & $\frac{d(x^*)}{ d(x^n)} < 1$ & $\frac{u(x^*)}{ u(x)} > 1$&
$\frac{d(x^*)}{d(x)} < 1$ & $\frac{u(x^*)}{u(x^n)} >1$ & $\frac{d(x^*)}{ d(x^n)}<1$\\
\hline
FGSM & $92.2$\% & $95.6$\% & $79.5$\% & $90.0$\% & $74.7$\% & $70.1$\% & $68.2$\% & $69.6$\% \\
BIM-A & $99.2$\% & $98.0$\% & $99.5$\% & $98.7$\% & $83.4$\% & $76.4$\% & $83.3$\% & $76.8$\%\\
BIM-B & $60.7$\% & $90.5$\% & $35.6$\% & $86.7$\% & $4.0$\% & $98.8$\% & $3.6$\% & $99.1$\% \\
JSMA & $98.7$\% & $98.5$\% & $97.5$\% & $96.5$\% & $93.5$\% & $91.5$\% & $87.4$\% & $89.6$\% \\
C\&W & $98.5$\% & $98.4$\% & $96.6$\% & $97.5$\% & $92.9$\% & $92.4$\% & $88.23$\% & $90.4$\% \\
\hline
\end{tabular}
\end{center}
\vspace*{-0.8em}
\caption{The uncertainty of an adversarial sample is typically larger than that of
its noisy and normal counterparts, and the density estimate is typically smaller.
$x^*$ indicates an adversarial sample, $x$ is a regular sample and $x^n$ a noisy
sample.}
\label{tab:values}
\end{table*}

We implement each of the four attacks (FGSM, BIM, JSMA, and C\&W) described in 
Section \ref{sec:attacks} in TensorFlow, using the {\tt cleverhans} library for FGSM
and JSMA \cite{cleverhans}. For the BIM attack, we implement two versions: BIM-A, 
which stops iterating as soon as miclassification is achieved (`at the decision boundary'),
and BIM-B, which runs for a fixed number of iterations that is well beyond the 
average misclassification point (`beyond the decision boundary'). For each attack type, we also
craft an equal number of noisy test samples as a benchmark. For FGSM and BIM,
these are crafted by adding Guassian noise to each pixel with a scale set so
that the mean $L_2$-norm of the perturbation matches that of the adversarial samples.
For JSMA and C\&W, which flip pixels to their min or max values, these are crafted by observing the number of pixels that were
altered in the adversarial case and flipping an equal number of pixels
randomly. Details about model accuracies on the adversarial sets and average perturbation
sizes are provided in Table \ref{tab:attack_details}. Some examples of normal, noisy
and adversarial samples are displayed in Figure \ref{fig:normal_noisy_adv}.

\subsection{Network setup}
\label{sec:network_setup}

Here, we briefly describe the models used for each dataset and their accuracies
on normal and noisy test samples.

\begin{itemize}
\item {\bf MNIST:} We use the LeNet \cite{lecun1989backpropagation} convnet architecture with 
a dropout rate of 0.5 after last pooling layer and after the inner-product
layer. This model reports $98.7\%$ accuracy on normal samples and $97.2\%$ accuracy 
on noisy samples.

\item {\bf SVHN:} We use the LeNet architecture with an extra intermediate inner-product 
layer to assist with higher dimensionality. We used a dropout rate of 0.5 after the 
last pooling layer and after each inner-product layer. This model reports $92.2\%$ 
accuracy on normal samples and $79.2\%$ accuracy on noisy samples.

\item {\bf CIFAR-10:} We use a deep 12-layer convnet with a dropout rate of 0.5 applied 
after the last pooling layer and after each of the 2 inner-product layers. This model 
reports $82.6\%$ accuracy on normal samples and $79.2\%$ accuracy on noisy samples.
\end{itemize}

Training was done using the Adadelta optimizer with cross-entropy loss and a batch size of 256.

\subsection{Feature Values}

When we generate adversarial samples, the uncertainty typically grows larger than
the original sample, and the density estimate typically grows smaller.  This makes
sense: the adversarial sample is likely to be in a region of higher uncertainty,
lower density estimates, or both. In addition, the change is far more
pronounced than if we simply perturb the sample with random noise.

In order to demonstrate this phenomenon, we generate adversarial samples and
randomly perturbed (noisy) samples from the test data points for MNIST and 
CIFAR-10. For each attack, we calculate the percentage of points with 
higher uncertainty values than the corresponding
original unperturbed samples, and the percentage of points with lower density
estimates than the corresponding original unperturbed samples.  The results are
shown in Table \ref{tab:values}.

We can see clearly that uncertainty is generally increased when adversarial
samples are generated, and density estimates are generally decreased. 
These results suggest that our two features are reliable indicators of adversarial 
samples. Therefore, we next move on to the task of detecting adversarial samples.

\subsection{Adversarial Sample Classifiers}

\begin{table*}[t!]
\begin{center}
\begin{tabular}{|c|c|c|c|c|c|c|}
\hline
{\bf Dataset} & {\bf FGSM} & {\bf BIM-A} & {\bf BIM-B} & {\bf JSMA} & {\bf C\&W} & {\bf Overall}\\
\hline
MNIST & $90.57$\% & $97.23$\% & $82.06$\% & $98.13$\% & $97.94$\% & $92.59$\%\\
CIFAR-10 & $72.23$\% & $81.05$\% & $95.41$\% & $91.52$\% & $92.17$\% & $85.54$\%\\
SVHN & $89.04$\% & $82.12$\%& $99.91$\% & $91.34$\% & $92.82$\% & $90.20$\%\\
\hline
\end{tabular}
\end{center}
\vspace*{-0.5em}
\caption{ROC-AUC measures for each dataset and each attack for the logistic
regression classifier ({\tt combined}).}
\label{tab:rocaucs}
\end{table*}

\begin{figure*}[t!]
\begin{center}
\begin{subfigure}[b]{0.33\textwidth}
\begin{center}
\includegraphics[width=\textwidth]{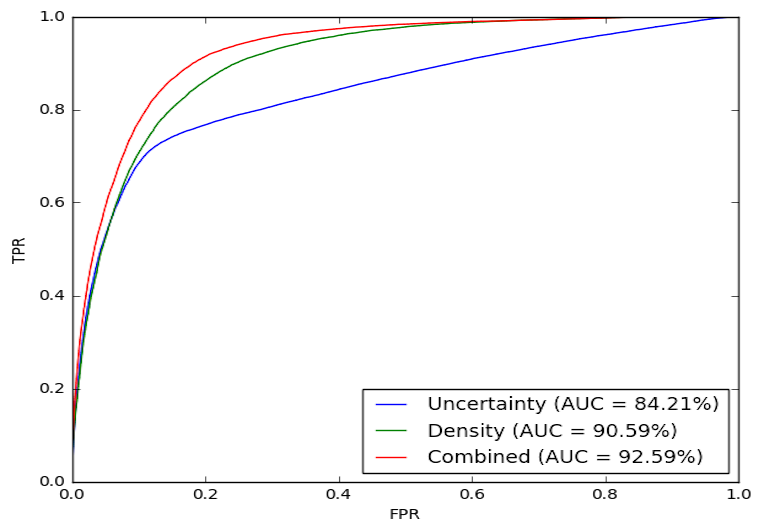}
\end{center}
\vspace*{-1.0em}
\caption{MNIST dataset.}
\label{fig:rocs:mnist}
\end{subfigure}
\begin{subfigure}[b]{0.33\textwidth}
\begin{center}
\includegraphics[width=\textwidth]{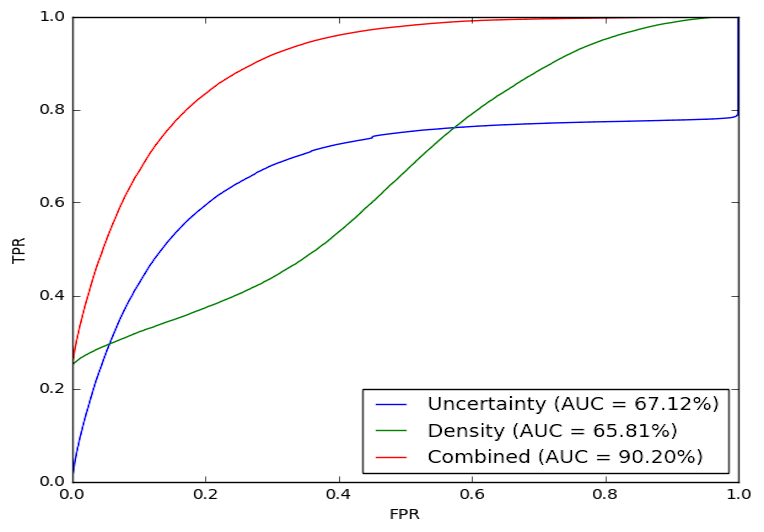}
\end{center}
\vspace*{-1.0em}
\caption{SVHN dataset.}
\label{fig:rocs:svhn}
\end{subfigure}
\begin{subfigure}[b]{0.33\textwidth}
\begin{center}
\includegraphics[width=\textwidth]{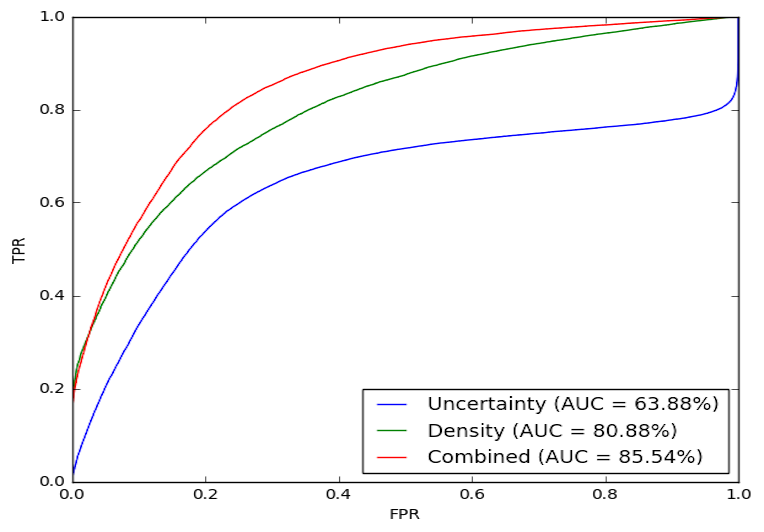}
\end{center}
\vspace*{-1.0em}
\caption{CIFAR-10 dataset.}
\label{fig:rocs:cifar}
\end{subfigure}
\end{center}
\vspace*{-1.0em}
\caption{ROCS for the different classifier types. The blue line indicates {\tt uncertainty}, green line 
{\tt density}, and red line {\tt combined}. The negative class consists of both normal and noisy samples.}
\label{fig:rocs}
\end{figure*}

To evaluate our adversarial detection method, we aggregate all adversarial samples of different types 
into a unified set, and do the same with the noisy and normal samples. For each dataset and attack, 
we have built three binary classifiers:

\begin{itemize}
  \item {\tt uncertainty}: this classifier simply thresholds on the uncertainty value of
a sample.
  \item {\tt density}: this classifier simply thresholds on the negative log kernel
  density of a sample.
  \item {\tt combined}: this is a simple logistic regression classifier with two
features as input: the uncertainty and the density estimate.
\end{itemize}

These detection models are used to distinguish adversarial samples--the positive class--from 
normal and noisy samples, which jointly constitue the negative class. The logistic regression model 
is trained by generating adversarial samples for every correctly-classified training point 
using each of the four adversarial attacks, and then using the uncertainty values 
and density estimates for the original and adversarial samples as a labeled training set. 
The two features are z-scored before training.

\begin{figure}[b!]
\begin{center}
\includegraphics[width=0.45\textwidth]{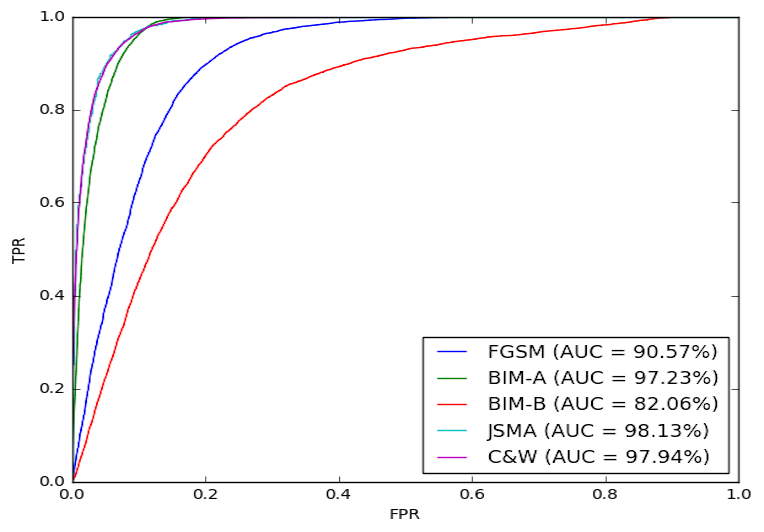}
\end{center}
\vspace*{-1.5em}
\caption{ROC results per adversarial attack for {\it combined} classifier on MNIST.}
\label{fig:rocs:mnist_separate}
\vspace*{-0.5em}
\end{figure}

Because these are all threshold-based classifiers, we may generate an ROC for
each method. Figure \ref{fig:rocs} shows ROCs for each classifier with a couple of datasets. 
We see that the performance of the {\tt combined} classifier is better than
either the {\tt uncertainty} or {\tt density} classifiers, demonstrating that
each feature is able to detect different qualities of adversarial features.
Further, the ROCs demonstrate that the uncertainty and density estimates are
effective indicators that can be used to detect if a sample is adversarial.
Figure \ref{fig:rocs:mnist_separate} shows the ROCs for each individual attack;
the combined classifier is able to most easily handle the JSMA, BIM-A and C\&W
attacks.

In Table \ref{tab:rocaucs}, the ROC-AUC measures are shown, for each of the
three classifiers, on each dataset, for each attack. The performance is quite good, suggesting
that the combined classifier is able to effectively detect adversarial samples
from a wide range of attacks on a wide range of datasets.

\section{Conclusions}
\label{sec:conclusion}

We have shown that adversarial samples crafted to fool DNNs can be effectively detected with two new
features: kernel density estimates in the subspace of the last hidden
layer, and Bayesian neural network uncertainty estimates. These two features handle complementary situations, and can be combined as an
effective defense mechanism against adversarial samples. Our results report
that we can, in some cases, obtain an ROC-AUC for an adversarial sample detector
of up to 90\% or more when both normal and noisy samples constitute the negative class. 
The performance is good on a wide variety of attacks and a range of image datasets.

In our work here, we have only considered convolutional neural networks.  However, we
believe that this approach can be extended to other neural network architectures
as well. \citet{GalRecurrent2015} showed that the idea of dropout as a Bayesian
approximation could be applied to RNNs as well, allowing for robust
uncertainty estimation. In future work, we aim to apply our features to RNNs
and other network architectures.

\section*{Acknowledgements}

We thank Nikolaos Vasiloglou and Nicolas Papernot for useful discussions that
helped to shape the direction of this paper. We also thank Symantec Corporation
for providing us with the resources used to conduct this research.


\bibliography{citations}
\bibliographystyle{icml2017}

\end{document}